\documentclass[runningheads]{llncs}

\usepackage{eccv}

\usepackage{eccvabbrv}
\usepackage{multirow}
\usepackage{xcolor}
\usepackage{graphicx}
\usepackage{booktabs}
\usepackage{bbding}
\usepackage{float}
\usepackage{wrapfig}
\definecolor{dgreen}{rgb}{0.0,0.6,0.0}
\usepackage{colortbl}  
\usepackage{xcolor}
\usepackage{array}
\usepackage[accsupp]{axessibility}

\usepackage{multirow}

\usepackage[pagebackref,breaklinks,colorlinks,citecolor=eccvblue]{hyperref}

\usepackage{orcidlink}

\begin{document}

\title{LaserHuman: Language-guided Scene-aware Human Motion Generation \\in Free Environment} 
\makeatletter
\let\@oldmaketitle\@maketitle
\renewcommand{\@maketitle}{
   \@oldmaketitle
 \begin{center}
    \vspace{-1mm}
\includegraphics[width=1.0\linewidth]{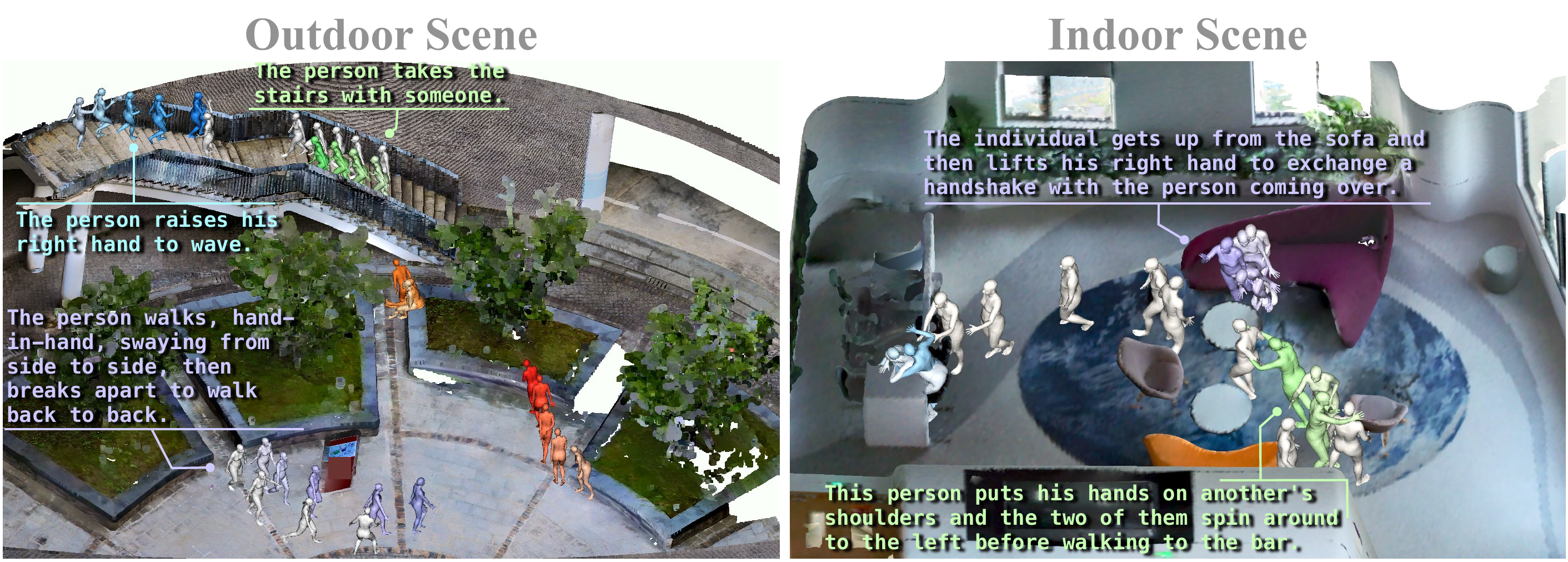}
 \end{center}
 \vspace{-1ex}
 \refstepcounter{figure}\normalfont Figure~\thefigure. 
  LaserHuman consists of large-scale sequences of rich human motions and abundant human interactions captured in various real scenarios with free-form language descriptions, providing valuable data for conditioned human motion generation. We demonstrate two scenarios, where colored humans are annotated targets and white humans are interacted humans in the dynamic scene. The human mesh color from light to dark represents an increase in timing.
  \label{fig:teaser}
  \newline
  }

\titlerunning{LaserHuman}

\renewcommand{\thefootnote}{\dag}

\author{Peishan Cong\inst{1}
\and
Ziyi Wang\inst{1}\and
Zhiyang Dou\inst{2,5}\and
Yiming Ren \inst{1}\and
Wei Yin \inst{3}\and
Kai Cheng \inst{4}\and
Yujing Sun \inst{5}\and
Xiaoxiao Long  \inst{5}\and
Xinge Zhu \inst{6}\and
Yuexin Ma \inst{1,}\textsuperscript{\dag}}
\footnotetext{Corresponding authors}

\authorrunning{P. Cong et al.}

\institute{ShanghaiTech University \and
University of Pennsylvania \and
University of Adelaide \and
University of Science and Technology of China \and
The University of Hong Kong \and
The Chinese University of Hong Kong \\
\email{\{congpsh,mayuexin\}@shanghaitech.edu.cn}\\ 
\url{https://github.com/4DVLab/LaserHuman}}

\maketitle
\begin{abstract}

Language-guided scene-aware human motion generation has great significance for entertainment and robotics. In response to the limitations of existing datasets, we introduce \textit{LaserHuman}, a pioneering dataset engineered to revolutionize Scene-Text-to-Motion research. LaserHuman stands out with its inclusion of genuine human motions within 3D environments, unbounded free-form natural language descriptions, a blend of indoor and outdoor scenarios, and dynamic, ever-changing scenes. 
Diverse modalities of capture data and rich annotations present great opportunities for the research of conditional motion generation, and can also facilitate the development of real-life applications.
Moreover, to generate semantically consistent and physically plausible human motions, we propose a multi-conditional diffusion model, which is simple but effective, achieving state-of-the-art performance on existing datasets. \textbf{Our dataset and code will be released soon.}

\end{abstract}    
\section{Introduction}
\label{sec:intro}

Generating realistic human motions from natural language descriptions in 3D scenes (Scene-Text-to-Motion) is a challenging task spanning computer vision, computer graphics, natural language processing, and robotics. Its impact on downstream applications is twofold. In domains like simulation, animation, and VR/AR, it replaces frame-by-frame 3D modeling with language-driven data generation, improving data creation efficiency and meeting user-specific requirements. In the context of humanoid robots, it directly generates robot action sequences according to 3D scene data and language instructions, as depicted in Fig.~\ref{fig:teaser}. This revolutionary approach shifts from traditional modular systems involving perception, prediction, and planning towards an end-to-end generation paradigm, reducing error accumulation and redundant computation.

Previous research in human motion generation primarily concentrates on generating motions based on a single condition~\cite{zhu2023human}. Text-to-motion approaches~\cite{Guo_2022_CVPR_t2m,petrovich22temos, mdm2022human,chuan2022tm2t,ahuja2019language2pose,kim2022flame} aim to generate corresponding motions from action classes or natural language descriptions, with evaluations conducted on datasets such as KIT~\cite{plappert2016kit}, HumanML3D~\cite{Guo_2022_CVPR_t2m}, BABEL~\cite{punnakkal2021babel}, etc. While this has propelled the advancement of digital human applications, its applicability to robotics remains limited due to the absence of scene constraints. Scene-to-motion methods~\cite{scenediff}, on the other hand, strive to produce plausible human motions coherent with the contextual scene. These methods typically rely on datasets focusing on human-scene interaction, including PROX~\cite{hassan2019resolving}, GTA-IM~\cite{wang2021synthesizing}, CIRCLE~\cite{araujo2023circle}, and others, for evaluation. Although meaningful for simulation and automated navigation in robotics, they lack the ability to incorporate precise control over language instructions.

Recently, HUMANISE~\cite{wang2022humanise} introduced a platform for Scene-Text-to-Motion research, offering textual and scene-based conditions for human motion generation. However, it faces four critical limitations: synthesized human motions, templated language descriptions, indoor scenes, and static environments, hindering its broader application. Firstly, it relies on entirely synthesized motion data, compromising the authenticity and physical plausibility of human movements. Secondly, fixed templates and restricted action types limit language descriptions, failing to accommodate real-world applications' diverse language and actions. Thirdly, it only supports indoor scenes with limited activity space and terrain variety. Fourthly, its environments consist solely of static furniture, whereas real-life scenarios often involve digital humans or humanoid robots interacting with dynamic elements like moving humans or objects.

In order to address the data limitations for the research of Scene-Text-to-Motion methods and extend their applicability to a more comprehensive range of scenarios, we propose a new large-scale dataset, named \textbf{LaserHuman}, tailored for \underline{\textbf{La}}nguage-guided \underline{\textbf{s}}cen\underline{\textbf{e}}-awa\underline{\textbf{r}}e \underline{\textbf{Human}} motion generation in free environments. Distinguished from existing datasets, LaserHuman comprises \textbf{real human motions in 3D scenes, free-form neural language descriptions, a mix of indoor and outdoor scenarios, as well as both static and dynamic environments}, as depicted in Fig.~\ref{fig:teaser}. We utilize calibrated and synchronized multiple 128 beam LiDARs, RGB cameras, and wearable IMUs to capture the scene and human motions, and then assign annotators with different ages and genders to give detailed linguistic descriptions for motion sequences. Especially, our capture scenes are large-scale real-life unconstrained scenarios, where the largest one spans over 2000 $m^2$ and the target human is freely interacting with static or dynamic objects inside. In total, LaserHuman contains 11 diverse 3D scenes, 3,374 high-quality motion sequences, and 12,303 language descriptions.

The task of Scene-Text-to-Motion is extremely challenging, because it requires the model to generate natural and realistic human motions that are semantically consistent with language descriptions and physically plausible with the interacted 3D scenes. To tackle the issues, we propose a new solution for language-guided scene-aware human motion generation with a multi-conditional diffusion model. In particular, we design a simple but effective multi-condition fusion module to enhance the consistency of generated motions with both text instructions and 3D scenes.
Our new method for Scene-Text-to-Motion has been evaluated by extensive experiments on LaserHuman and HUMANISE, and all achieve state-of-the-art performance.

Our contribution can be summarized as follows:
\begin{enumerate}
    \item We propose a novel Scene-Text-to-Motion dataset containing diverse human motions, free-form language descriptions, and unconstrained real-life scenarios, which is significant for many downstream applications. 
    \item To generate semantically consistent and physically plausible motions, we develop a multi-conditional diffusion model for the Scene-Text-to-Motion task, which achieves SOTA performance on existing datasets.
    \item We provide comprehensive discussions on our benchmark and dataset, which 
    has potential to boost the development in related research fields.
    
\end{enumerate}


    

\section{Related Work}
\label{sec:related}
\subsection{Human Motion Datasets}
Human motion representation is essential for social-behavioral modeling and analysis and can further benefit many downstream applications.
With the development of human motion capture~\cite{mahmood2019amass,sim2002cmu,ParkJS2012}, many datasets~\cite{guzov2021human, dai2022hsc4d,dai2023sloper4d,yan2023cimi4d,lip,ren2024livehps} provide SMPL models with global positions in 3D scenes for facilitating the research on 3D pose estimation. However, these datasets focus on local pose capture or global trajectory prediction in scenes, lacking scene-aware motion diversities and interactions, and also lacking textual annotations, not suitable for motion generation tasks.
~\cite{Guo_2022_CVPR_t2m,plappert2016kit,punnakkal2021babel} provide human motion sequences with textual descriptions, including language or action labels, covering a wide range of activities in daily life. 
~\cite{zhang2022couch,xiao2023unified} considers the interaction with specific target objects.
Nevertheless, they are limited to the text-to-motion task and lack the absence of 3D environments. 

%
Several datasets~\cite{hassan2019resolving,savva2016pigraphs,wang2021synthesizing,hassan2021stochastic} pay attention to human-scene interaction (HSI). PROX ~\cite{hassan2019resolving} employs optimization methods with RGB-D to reconstruct human motions, though the pose-fitting results are suboptimal in certain cases. 
To enhance motion quality, ~\cite{araujo2023circle,wang2021synthesizing} gather data on synthesized human-scene interactions.
\cite{zhao2022compositional,wang2022humanise} incorporate the context of both scenes and language descriptions on specific actions. 
However, these efforts mainly expand upon indoor scene data ~\cite{dai2017scannet,hassan2019resolving}, which are limited to static environments with regular layouts. Furthermore, their templated language descriptions hinder broader applicability in real-life scenarios.
To tackle above issues,
we collect a large-scale real-life motion dataset in unconstrained scenarios with rich interactions and free-language descriptions.

\subsection{Conditional 3D Human Motion Generation}
Significant progress has been made in 3D Human Motion Generation conditioned on texts or 3D scenes. There are two main categories of methods recently, including
 conditional variational auto-encoder (cVAE) ~\cite{zhang2020generating,cai2021unified,petrovich2021action,petrovich22temos,Guo_2022_CVPR_t2m,jiang2024motiongpt} methods, and conditional diffusion-based methods~\cite{mdm2022human,mdmprior,latentmdm,motiondiffuse,scenediff,xu2023interdiff,zhang2023remodiffuse,tevet2022motionclip}. The former generates plausible motion sequence by a VAE architecture, which learns a latent space constrained on a Gaussian distribution. The latter learns the relation between motion and conditioned input, such as language or scene, by the diffusion sampling process. However, the above methods are only conditioned on a single modality, lacking the ability to generate plausible motion on both textual and scene-based conditions. 
 ~\cite{wang2022humanise,xuan2023narrator} 
 generate human-scene interactions under language descriptions with a simple fusion condition module on VAE architecture. These rely on the learned latent variables suffering from the posterior collapse problem and have limited generation diversity, which cannot used in general 3D large scale with complex scene and language conditions. In contrast, we propose a fusion strategy for scene-text-conditioned generation with a diffusion model. 
 

\section{Dataset}


Previous datasets for human motion generation are mainly focusing on single modality conditions, either language or scene.  Even though \cite{wang2022humanise,xuan2023narrator} incorporate multi-modal constraints, they are restricted to static indoor environments, utilizing templated textual descriptions, which is not applicable for daily-life applications.
We introduce LaserHuman, a large-scale dataset for real-life, unconstrained scenarios complemented with free-form language descriptions, collected in both static and dynamic scenarios. 
Detailed data acquisition and pre-processing, statistics, and characteristics of our datasets are in the following.

\subsection{Data Acquisition and Pre-processing}

\begin{figure*}[ht]
    \centering
    \includegraphics[width=0.98\columnwidth]{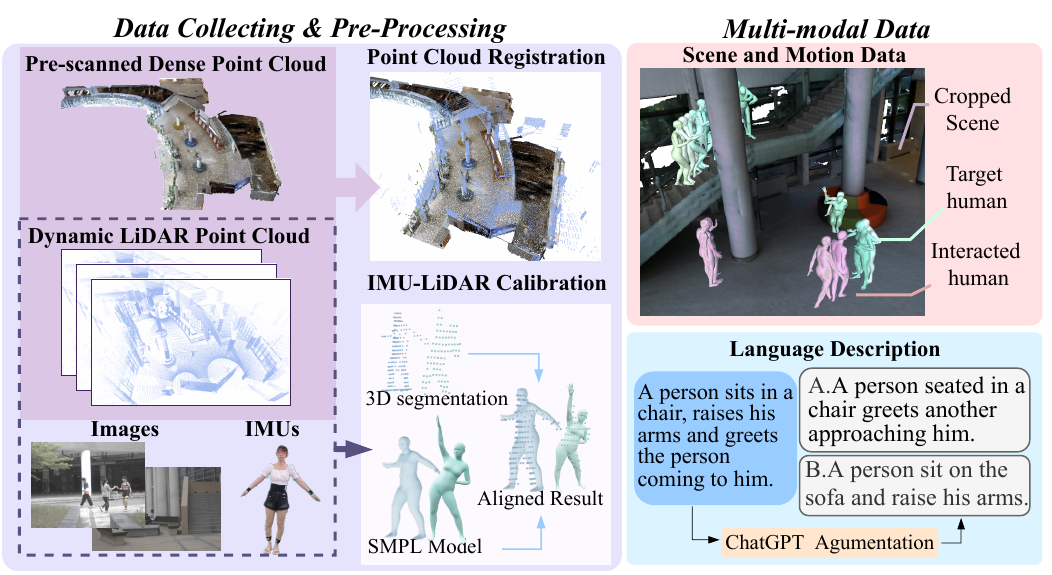
    }
    \caption{Overview of data collection and processing procedures.}
    \label{fig:collection}
    \vspace{-4ex}
\end{figure*}
\noindent\textbf{Data Collection.}
As shown in Fig.~\ref{fig:collection}, LaserHuman includes pre-scanned dense point cloud with Stereye Polar 1S. A 3D scene map is obtained by SLAM algorithms and the mesh is constructed by~\cite{ladicky2017point}. To collect dynamic scenes, we set up synchronized and calibrated multiple 128-beam LiDARs and RGB cameras in different views. 
For each sequence of motion, there is a target person wearing Noitom~\cite{noitom} devices with 17 IMUs and making interaction with the objects or other persons naturally. We involve up to three persons to make close interaction with the target person. For the diversity of motions, we collect motions of different target persons with different ages and genders.

\begin{figure*}[ht]
    \centering
    \includegraphics[width=\linewidth]{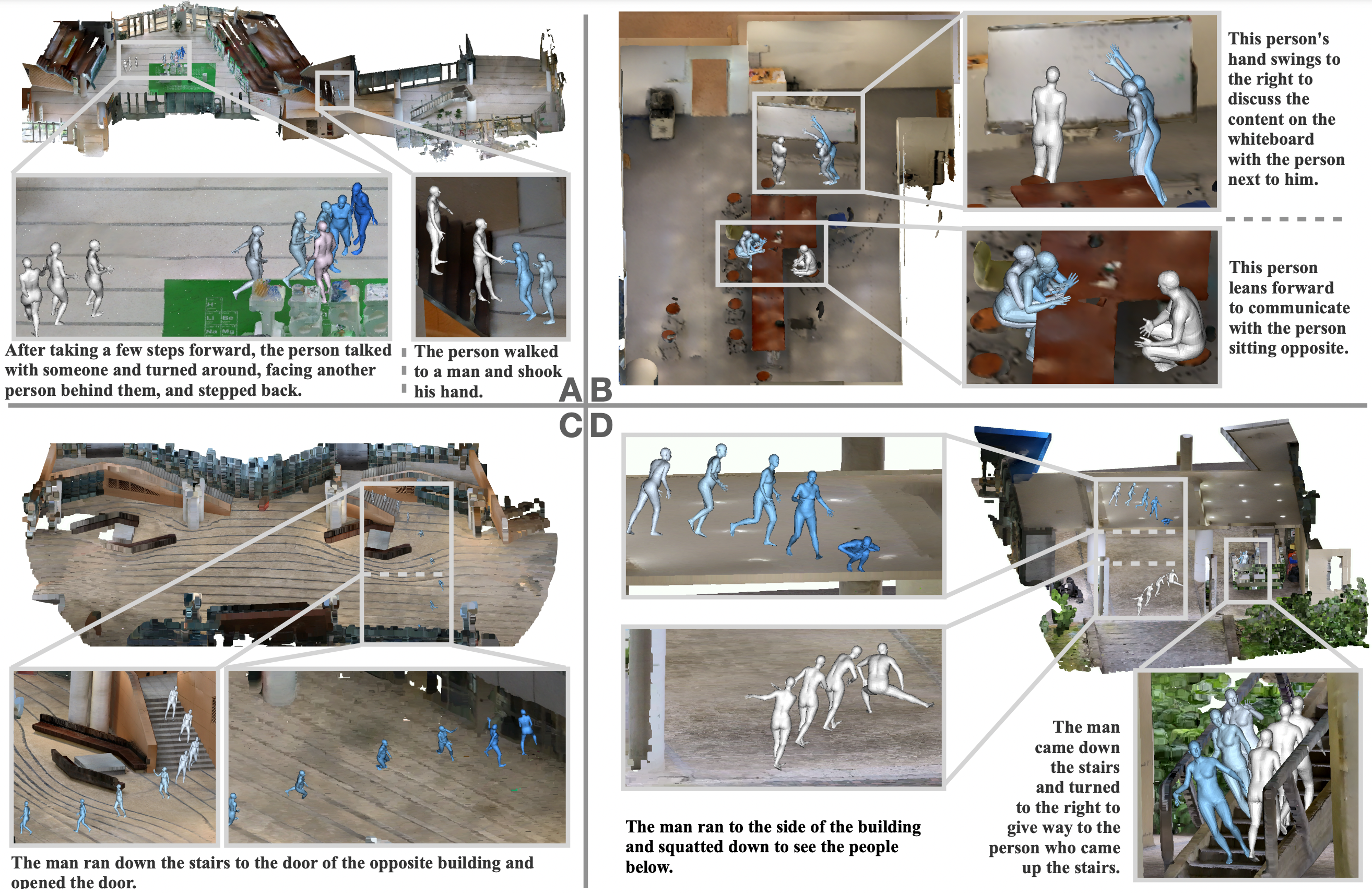}
    \caption{A gallery of LaserHuman, containing diverse scenarios, rich interactions, and abundant free-form language descriptions, where colored humans are annotated targets and white humans are interacted humans in the dynamic scene.}
    \label{fig:gallary}
\end{figure*}

\noindent\textbf{Data Pre-Processing.}
The pre-scanned scene maps are aligned with the dynamic point clouds collected by LiDAR by registration algorithm ~\cite{koide2021voxelized}. 
Since LiDAR can provide accurate depth information, we segment the human points in LiDAR and follow the optimization algorithm in \cite{dai2023sloper4d} and \cite{lip} to obtain the global translation $T$ in world coordinate. 
SMPL~\cite{loper2023smpl} is used to represent the human body motion $x = (T,\theta,\beta)$ , where $T,\theta \in \mathcal{R}^{J\times3},\beta$ are the translation parameter, body rotation parameters, and shape parameters. $J$ is the joint numbers (23 joints’ rotation relative to their parent joint and one root rotation). The aligning process is semi-automatic, we manually segment the person and adjust positions in some cases with significant occlusion. We have achieved the average Chamfer Distance of 4.5mm between the point cloud and the mesh, which indicates an accurate alignment. Noted we focus on the motion generation only for target person. The SMPL of other interacted participants is obtained by~\cite{ren2023lidar} and optimized from multi-view LiDAR point clouds. More processing details are in the appendix. 

\noindent\textbf{Multi-modal Data.}
LaserHuman provides multi-modal data including videos, 3D scene maps, dynamic LiDAR point clouds, and global human motions for the target person and other participants.
We further obtain the language annotation in flowing process. Motion data are segmented into clips, each clip will contain motion sequences ranging from 1 to 4 seconds. 
Two different people will give free-form linguistic descriptions for each sequence, respectively. Annotators are from various ages and genders, enriching the description diversity.
These descriptions detail the person's movements and the interaction with the dynamic environment. We employ ChatGPT for text augmentation by expanding the number and diversity of texts.
To guarantee the correctness and uniqueness of the linguistic description for targets, two people take turns checking the motion-text consistency and quality.

\begin{table*}[t]
\centering
\caption{Comparison with datasets for scene-aware human motion generation, where ``Num'' represents number, ``PC'' is point cloud. For the scene scale, 30 $m^2$ is approximated for single indoor room. }
\vspace{-2ex}
\resizebox{\linewidth}{!}{
\begin{tabular}{c|c|c|cc|ccc|ccccc}
\hline
\rowcolor{blue!5} \textbf{Data Source}        & \textbf{Dataset} & \textbf{Modality}  & \multicolumn{2}{c|}{\textbf{Motions}} & \multicolumn{3}{c|}{\textbf{Text}}        & \multicolumn{5}{c}{\textbf{Scene}}                     \\ \cline{4-13} \rowcolor{blue!5}
\multicolumn{1}{l|}{}         &  &                     & Duration   & Seqences   & Num & Form     & Length & Num & Scale      & Indoor & Outdoor & Dynamic \\ \hline
\multirow{2}{*}{Synthesized} & CIRCLE \cite{araujo2023circle}              & SMPL,RGB-D                   & 10h           & 7k           &    -          &      -    &  -      & 9   & $30\sim100 m^2$     & \textcolor{green}{\Checkmark}    & \textcolor{red}{\XSolidBrush}&\textcolor{red}{\XSolidBrush}        \\ \cline{2-13} 
& HUMANISE~\cite{wang2022humanise}           & SMPL,RGB-D,text              & 10h           & 19.6k        & 19.6k        & template & 7.6    & 643 & $10\sim30 m^2$     & \textcolor{green}{\Checkmark}    & \textcolor{red}{\XSolidBrush}&\textcolor{red}{\XSolidBrush}          \\ \hline
\multirow{2}{*}{Real-world}  & PROX~\cite{hassan2019resolving}                 & SMPL,RGB-D                   & 1h            & 28k          &    -          &     -     &  -      & 12  & $10\sim30 m^2$ & \textcolor{green}{\Checkmark}      & \textcolor{red}{\XSolidBrush}&\textcolor{red}{\XSolidBrush}       \\ \cline{2-13} 
& LaserHuman           & SMPL,RGB-D,text,PC,Scene-Map & 3h            & 3.5k         & 12.3k        & free     & 28.5   & 11  & $50\sim2000 m^2$    &     \textcolor{green}{\Checkmark}&\textcolor{green}{\Checkmark}&\textcolor{green}{\Checkmark}   \\ \hline
\end{tabular}}
 \label{tab:data}
  \vspace{-2ex}
\end{table*}

\subsection{Statistics}
\label{sec:statistics}
We compare with other open datasets for \textbf{scene-aware motion generation} with detailed statistics in Table.~\ref{tab:data}. LaserHuman consists of 3,374 motion sequences with a duration from 1 to 4 seconds, with 4,495 free-form text annotations, the average length of the sentence description is 28.5 words, which is four times longer than existing datasets. After text augmentation, the total count of language descriptions is expanded to 12,303. The total point cloud frames number is 92,955 with a collection time of more than 3 hours.

LaserHuman provides 11 pre-scanned large scenes with the largest spanning over 2000 $m^2$, which consist of 5 outdoor areas, 2 relatively small indoor, and 4 complexly designed first-floor halls. 
The scenarios include diverse scenes such as corridors, benches, rooms with sofas and tables, staircases, and doorways. 
To facilitate the use of generative models, we crop pre-scanned dense point clouds of static scenes according to each motion sequence into areas exceeding 150 $m^2$. 
\textbf{Accordingly, we prepare three kinds of scene data to boost follow-up studies}: the first one is the integration of cropped scene map and LiDAR point cloud sequences of interactive participants; the second one is comprised of cropped scene map and SMPL sequences of interactive participants; and the third one is only the cropped scene map. The former two types of scene data can facilitate the research of dynamic scene-conditioned motion generation with different input modalities, while the latter one focuses on static scene-conditioned motion generation.

\subsection{Characteristics}

Fig.~\ref{fig:gallary} shows several scenarios in our dataset.
Compared with existing open datasets for scene-aware motion generation, LaserHuman has four main novel characteristics as follows.

\noindent\textbf{Real human motions in scenes.} 
~\cite{wang2022humanise,araujo2023circle} refine the motion data in a synthesis fitting process by adjusting the global translation and rotation to ensure compatibility with 3D scenes. However, it, to some extent, compromises the authenticity and physical plausibility of human movements. Our dataset is collected in real human-scene interaction scenarios, maintaining rational physics and natural interaction in the real world.

\noindent\textbf{Diverse human motion categories.}
HUMANISE\cite{wang2022humanise} is the only open motion dataset with both text and scenes but limits itself to 4 basic human postures: sit, stand, walk, and lying down, our dataset delves into more intricate details. PROX\cite{hassan2019resolving} and CIRCLE\cite{araujo2023circle} are also confined to static indoor scenes and lack detailed movement captures.
Our dataset includes detailed depictions of interaction with static objects and terrains, complex sports activities and person-to-person interactions(eg. shaking hands, touching head, hugging, and greeting), which are notably absent in these other datasets. Diverse motion visualization are shown in Fig. \ref{fig:gallary} and supplementary materials. 

\noindent\textbf{Diverse interactive environments.}
Previous scene-aware motion generation datasets focus on static indoor environments with flat ground. However, we are living in a dynamic world with moving and changing objects. Generating realistic human motions in dynamic scenarios is crucial for simulation, animation, and the realm of humanoid robots. As Fig. \ref{fig:gallary} shows, our dataset concerns the interaction with diverse environments including both static objects and moving individuals. The responding behavior to dynamically moving objects, such as obstacle avoidance and body contact, is a crucial factor to be considered in the generation of human motions. LaserHuman provides the opportunity for the research of dynamic scene-constrained motion generation.
Moreover, our dataset contains both indoor and outdoor scenes with various terrains, such as stairs, bringing challenges for physical-plausible motion generation.

\noindent\textbf{Free-form neural language descriptions.}
Previous scene-aware motion generation datasets~\cite{wang2022humanise,xuan2023narrator} adopt constrained templates for text descriptions, primarily limiting to a few number of specific short-term actions, likes sit or walk.
In contrast, our dataset contains free-form neural language descriptions for short-term or long-term motions or trajectories with action details, aiming to enhance the applicability and realism of generating motions in real-world scenarios.






\section{Method}

\begin{wrapfigure}{r}{.5\linewidth}
\centering\includegraphics[width=0.5\columnwidth]{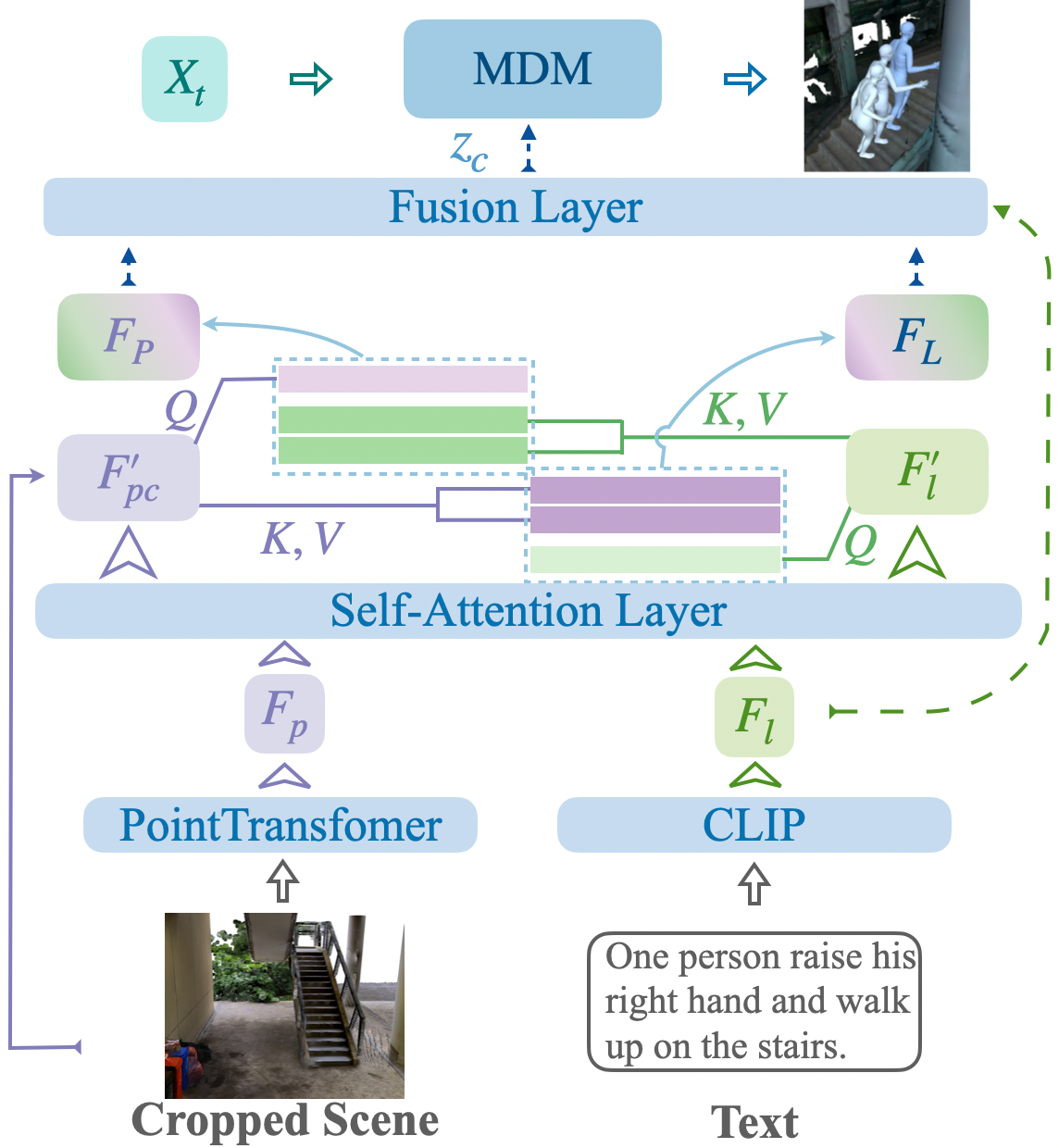}
    \caption{The pipeline of our generative model, which is applicable for language-guided scene-aware human motion generation. We demonstrate details of the multi-condition fusion module.}
    \label{fig:framework}
    \vspace{-3ex}
\end{wrapfigure}

Our goal is to generate human motion sequences that are semantically consistent with textual descriptions and physically plausible with 3D scenes. The overall architecture is shown in Fig. \ref{fig:framework}. In particular, we present a multi-condition fusion module in our diffusion-based generative model for effectively integrating both scene and textual contexts.
\subsection{Multi-condition Fusion Module}
Our method is based on the first kind of input introduced in Section.~\ref{sec:statistics}, where the scene data is the registered point cloud consisting of static scene map and dynamic point cloud sequences. We have a simplification of the large scenario, with the cropped point clouds  as input and normalize the scene center to (0,0), which also serves as the start position of the motion sequence.
The condition module takes the scene data and a textual description as input and outputs a joint conditional embedding. 
For the initial feature extraction for two conditions, point Transformer~\cite{pointtransformer} is employed to extract the scene features $F_p$ from the original point feature $f_p \in \mathbf{R}^{N_p \times D}$, where $N_p$ is the downsampled point cloud number and $D=6 $ representing the $(x,y,z,r,g,b)$ features. For language description, a pre-trained CLIP~\cite{clip} is applied to extract the language feature $F_l$. 

For text-conditioned or scene-conditioned motion generation tasks, the generated motion is required to be semantically correct under motion instruction in text, or is demanded to be physically plausible in scene. For scene-text-conditioned motion generation, besides these constraints, the model also needs to understand description of scene semantics in language and generate realistic interactions with the scene under multi-modal conditions.
To our knowledge, there are two related works for scene-text multi-conditioned    generation. \cite{wang2022humanise} takes a simple concatenate and self-attention strategy, which overlook the intricate interplay between modalities. \cite{vuong2023language} adopts a dual scene-queried approach for scene synthesis under multi-condition. It over-emphasizes the scene information while neglecting text instruction, which is not suitable for motion generation.

Text feature contains the specific action and motion direction condition, which constrain the locomotion. Meanwhile, due to the language constraint, the target should interact at specific location in the scene. Point cloud feature contains rich geometry and dynamic information, which is important for final text-matched plausible interaction with objects in environment. 
We design a simple but effective fusion method to fuse two modalities of conditions and then guide the diffusion process.
We first utilize a self-attention layer and we get the enhanced feature $F_{p}'$ and $F_l'$. $F_p'$ and original position feature $f_p$ are concatenated and forwarded with a fully-connected layer, $F_{pc}' = FC(F_p'\oplus f_p))$. A parallel cross-attention mechanism is then employed to learn the feature interaction between different modalities. Features are utilized as queries to mutually enhance the counterpart. The whole process is represented as:
\begin{equation}
\begin{aligned}
F_L &= LN(FFN(\text{CA}(F_l', F_{pc}', F_{pc}') + F_l'),\\
F_P &= LN(FFN(\text{CA}(F_{pc}', F_l', F_l')+F_{pc}').
\end{aligned}
\end{equation}
CA is the cross-attention operation, LN is layer normalization and FFN is the feed-forward neural network ~\cite{vaswani2017attention}. 
To preserve the original fine-grained text features, it is concerted with the fusion feature together to obtain the final condition feature,
$z_c = FC(F_l'\oplus F_P \oplus F_L).$

\subsection{Learning and Sampling}
We represent human motion as a sequence of
poses $\textbf{x} = \{x_n\}^{N_m}$, where $x_n$ is the single pose formed as the representation in HumanML3D~\cite{Guo_2022_CVPR_t2m}, $N_m$ is the length of the motion sequence.
The diffusion model~\cite{ho2020denoising} produces the latent $x_t$ by adding Gaussian noise on the motion sequence $x_0$, and generates the motion from $x_t$, a noise step $t$ and scene-text condition feature $z_c$. 
T Gaussian noising steps are modeled by the stochastic process,
\begin{equation}
    \begin{aligned}
q\left(\mathbf{x}_t \mid \mathbf{x}_{t-1}\right) &=\mathcal{N}\left(\mathbf{x}_t ; \sqrt{1-\beta_t} \mathbf{x}_{t-1}, \beta_t \mathbf{I}\right),
\\
\mathbf{x}_t&=\sqrt{\bar{\alpha}_t} \mathbf{x}_0+\sqrt{1-\bar{\alpha}_t}\epsilon,
    \end{aligned}
\end{equation}
where $\beta_t \in (0,1)$ is the variance of Gaussian noise. $\overline{\alpha_t}=\prod_{m=0}^t \alpha_m$ and $\alpha_t=1-\beta_t, \epsilon \sim \mathcal{N}(0, \mathbf{I})$.

We follow the denoising diffusion model MDM\cite{mdm2022human}
and feed $t,\mathbf{x}_t,z_c$ into a transformer model and predict the 
clean motion $\hat{x_0}=\mathcal{M}\left(\mathbf{x}_t, t, z_c\right)$. 
\begin{equation}
    L_{\text{motion}}=E_{t, \mathbf{x}_0}\left[\left|\mathbf{x}_t-\mathcal{M}\left(\mathbf{x}_t, t, z_c\right)\right|\right].
\end{equation}
$L_{\text{motion}}$ and the common geometric losses in \cite{mdm2022human} are applied to encourage natural and coherent motion.
During the sampling process, the clean motion $\hat{\mathbf{x}_0}$ is predicted and noised back to $\mathbf{x}_{t-1}$ and repeat T times in an iterative manner.

\section{Experiments}
To evaluate the model performance, we conduct a comparison with scene-text-to-motion methods on LaserHuman and HUMANISE\cite{wang2022humanise} and demonstrate a detailed comparison of various fusion strategies. More analysis and discussion on potential research directions are also provided.

\subsection{Evaluation Metrics}
We mainly follow the metrics in \cite{scenediff,mdm2022human}.
Non-collision score and contact score evaluate the physical plausibility of interacting with the scene, 
measuring whether a generated motion collides with other objects the score that body contact with the scene at a distance below a threshold, respectively. The diversities are measured by global translation, generated SMPL parameters, and the marker-based body-mesh representation in Average Pairwise Distance (APD-t/p/m) and standard deviation (std-t/p/m). FID measures the distance of the generated motion distribution to the ground truth distribution in latent space. R-score measures the text and motion matching accuracy. 
For general quality by human evaluations, we conduct a user study to measure the plausibility of generated motions. We randomly select 100 sequences from test scenes and instructed ten participants evaluate each sequence range from 1 to 5, with 5 denoting the highest quality. Participants were instructed to evaluate the generated results on four criteria: diversity of generation, scene consistency, text alignment, and smoothness. The overall plausibility score (denotes as p-score) for each result is calculated as the average of four aspects ratings. 
\begin{table*}[t]
\caption{Quantitative results of human motion generation on LaserHuman.}
\vspace{-2ex}
\resizebox{\linewidth}{!}{
\begin{tabular}{c|c|cccccccccc}
\hline
\rowcolor{blue!5}&     Methods       &non-collision$\uparrow$ & contact$\uparrow$ & APD (t)$\uparrow$ & std (t)$\uparrow$ & APD (p)$\uparrow$ & std (p) $\uparrow$
              & APD (m)$\uparrow$ & std (m)$\uparrow$
              & FID $\downarrow$& R-score$\uparrow$      \\ \hline
\multirow{2}{*}{text-only} & MDM \cite{mdm2022human}       & 0.853          & 0.512          & 1.116          & 0.347          & \textbf{12.969} & 0.588          & 4.130           & 0.109          & 1.014          & 0.313          \\ \cline{2-12} 
& MLD \cite{mld}       & 0.840          & 0.522          & 0.851          & 0.240        & 12.667          & 0.566          & 3.905          & 0.101          & 10.067         & 0.089          \\ \hline
scene-only                 & SceneDiff\cite{scenediff} & \textbf{0.999}          & 0.426          & 0.513          & 0.152          & 2.886           & 0.103          & 2.940           & 0.097          & 1.818          & 0.208          \\ \hline
\multirow{3}{*}{fusion}    & cVAE\cite{wang2022humanise}        & 0.998 & 0.459 & 0.756 & 0.264 & 0.151  & 0.124          & \textbf{7.018} & \textbf{0.294} & 1.595          & 0.195          \\ \cline{2-12} 
& sd-Text    & \textbf{0.999}          & 0.455          & 0.674          & 0.232          & 2.704           & 0.105          & 2.557          & 0.097          & 2.010           & 0.210           \\ \cline{2-12} 
& ours       & \textbf{0.999}          & \textbf{0.523} & \textbf{2.035} & \textbf{0.830}  & 12.903          & \textbf{0.847} & 4.117         & 0.152          & \textbf{0.987} & \textbf{0.326} \\ \hline
\end{tabular}}
 \vspace{-1ex}
\label{tab:r1}
\end{table*}

\begin{table}[t]
\caption{Detailed user study results of human motion generation on LaserHuman.}
\vspace{-2ex}
\resizebox{\linewidth}{!}{ 
\begin{tabular}{c|c|cccccc}
\hline
\rowcolor{blue!5}  & Methods & \multicolumn{1}{l}{diversity$\uparrow$} & \multicolumn{1}{l}{scene consistency$\uparrow$} & \multicolumn{1}{l}{text consistency$\uparrow$} & \multicolumn{1}{l}{smoothness$\uparrow$} & 
\multicolumn{1}{l}{total plausible score$\uparrow$} \\ \hline
text-only               & MDM\cite{mdm2022human}                  & \textbf{4.8}                           & 1.1                                   & 2.8                                  & \textbf{4.6}                           & 3.3                           \\ \hline
scene-only              & SceneDiff\cite{scenediff}           & 4.6                           & 2.2                                   & 0.8                                  & 3.8                            & 2.8                            \\ \hline
\multirow{3}{*}{fusion} & cVAE    \cite{wang2022humanise}              & 4.5                           & 2.7                                   & 1.1                                  & 4.5                            & 3.2      \\ \cline{2-7} 
& sd-Text    & 4.5                           & 3.8                                   & 2.1                                  & 3.6                            & 3.5                             \\ \cline{2-7} 
                        & ours                 & \textbf{4.8}                           & \textbf{3.9}                                   & \textbf{3.1}                                  & \textbf{4.6}                            & \textbf{4.1}                             \\ \hline
\end{tabular}}
\vspace{-3ex}
\label{tab:us}
\end{table}

\begin{figure*}[t]
    \centering
\includegraphics[width=1.00\columnwidth]{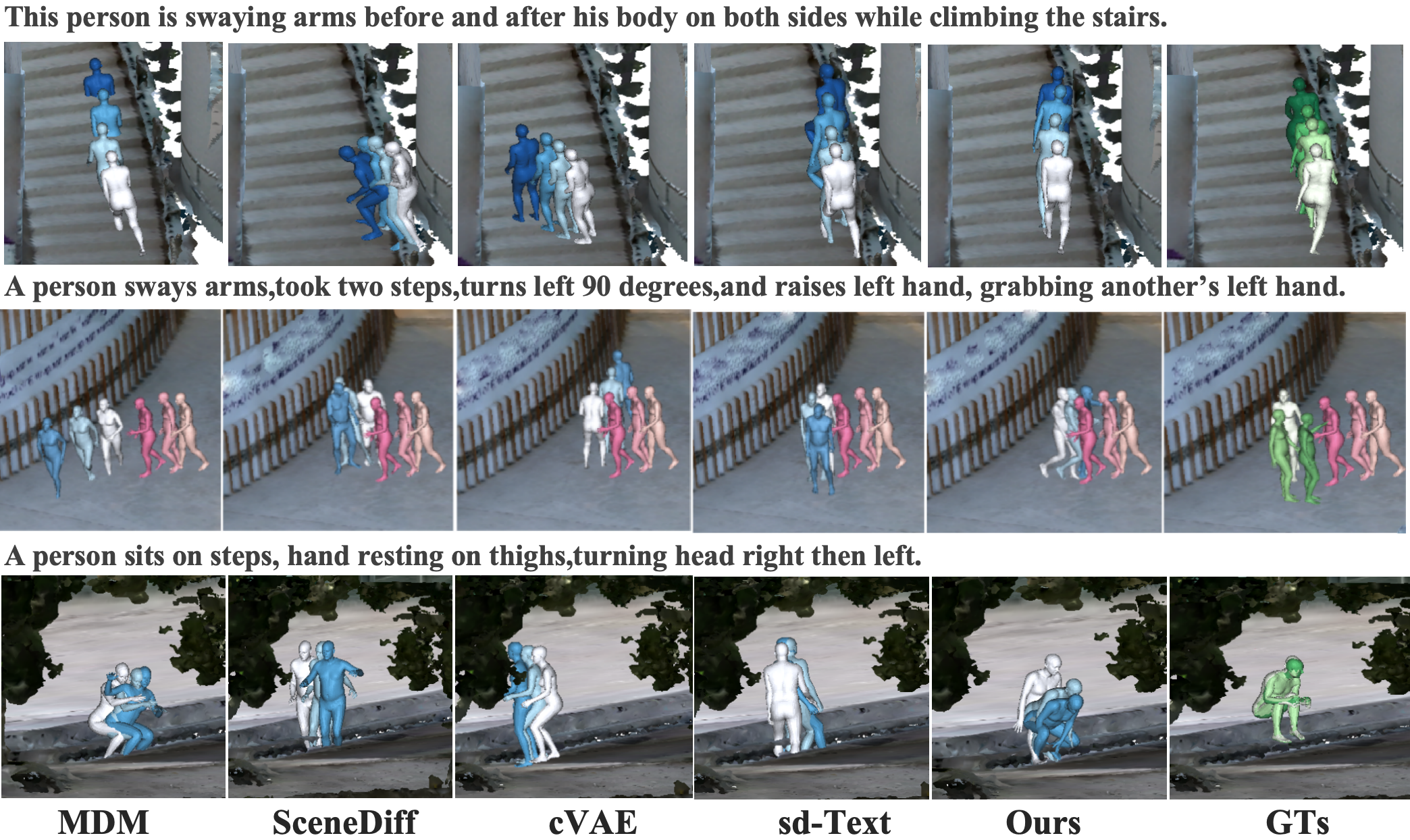}
\vspace{-4ex}
    \caption{Generation results on LaserHuman. The human mesh color from light to dark represents an increase in timing and pink human are corresponding interacted humans in the dynamic scene. }
    \label{fig:compare-result}
    \vspace{-1ex}
\end{figure*}

\begin{figure}[t]
    \centering
\includegraphics[width=0.98\columnwidth]{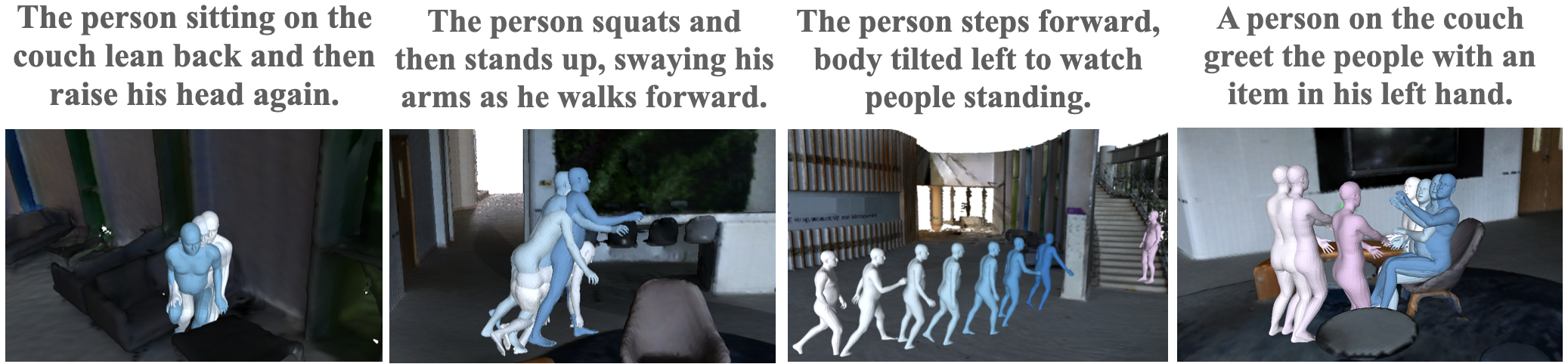}
  \vspace{-1ex}
    \caption{More generation results of our method on LaserHuman. }
    \label{fig:result}
     \vspace{-2ex}
\end{figure}

\subsection{Baselines}
Our goal is to generate human motion sequences that are semantically consistent with textual descriptions and plausible with 3D scenes, we compare with the state-of-the-art human motion generation methods that rely either single and multiple conditions. 
MDM\cite{mdm2022human} and MLD\cite{mld} are methods conditioned only on languages and SceneDiff\cite{scenediff} is scene-based conditioning. 
For multi-conditional motion generation task, 
\cite{wang2022humanise} is the only method. 
We also employ a straightforward fusion module\cite{wang2022humanise} on SceneDiff\cite{scenediff} and denote this modified variant termed sd-Text.

\subsection{Results}

The comparison result is illustrated in Table.~\ref{tab:r1} and detailed user studies results are shown in Table.~\ref{tab:us}.
Compared with conditioning on single-modality, we have a improvement on the metrics. The results of our detailed user study further underscore this advancement, particularly highlighting enhanced consistency within each respective modality. Conditioning only on the scene leads to a loss of textual constraints, resulting in scenarios that may occasionally appear nonsensical. 
Relying solely on languages presents a significant limitation in global spatial position, falling short in accurately representing the intricate spatial relationships inherent in dynamic scenes involving human. The generated motion may appear in the air shown in Fig.~\ref{fig:compare-result}. 
Compared with multi-conditional methods, our method has better results on contact scores, FID, R-score and translation diversity. cVAE shows less variance in parameters yet higher diversity in marker and translation metrics, indicating this modal generates a motion with limited variation on local pose while still having diverse global locations, leading to implausible sliding motions. sd-Text overlooks the intricate interplay between modalities, leading to a modality gap and poor performance. Benefiting from the interaction between two modalities, ours have superior results than others. Visual comparisons in Fig. \ref{fig:compare-result} further show our generation result is more plausible. 
However, achieving precise point-level contact remains a challenge, although our method shows greater consistency with language instructions for specific actions, such as ``grab hand".
Fig. \ref{fig:result} also shows our model generates human motions that are visually appealing and semantically consistent with the language description. 
We further conduct the experiment on HUMANISE~\cite{wang2022humanise} dataset in Table.~\ref{tab:r2}, which also provides both text and scene conditions for human motion generation. Our method achieves superior performance.

\begin{table*}[t]
\caption{Quantitative results of human motion generation on Humanise. }
\vspace{-3mm}
\resizebox{\linewidth}{!}{
\begin{tabular}{c|c|ccccccccccc}
\hline
\rowcolor{blue!5}   & Methods &non-collision$\uparrow$ & contact$\uparrow$ & APD (t)$\uparrow$ & std (t)$\uparrow$ & APD (p)$\uparrow$ & std (p) $\uparrow$
              & APD (m)$\uparrow$ & std (m)$\uparrow$
              & FID $\downarrow$& R-score$\uparrow$  & p-score $\uparrow$\\ \hline
text-only               & MDM\cite{mdm2022human}                   & 0.906         & 0.846   & 0.101      & 0.028      & 2.163      & 0.088      & 3.130       & \textbf{0.109}       & 0.392 & 0.121   &      3.9           \\ \hline
scene-only              & SceneDiff\cite{scenediff}             & \textbf{0.999}         & 0.794   & \textbf{1.060}      & \textbf{0.354}      & 2.522      & 0.086      & 2.228       & 0.087       & 2.525 & 0.189   &     3.1            \\ \hline
\multirow{3}{*}{fusion} & cVAE\cite{wang2022humanise}                   & 0.998         & 0.785   & 0.919      & 0.310      & 0.042      & 0.101      & 1.134       & 0.042       & 0.592 & 0.156   &         3.4       \\ \cline{2-13} 
                        & sd-Text               & \textbf{0.999}         & 0.839   & 0.907      & 0.232      & \textbf{2.704}      & 0.105      & 2.754       & 0.107       & 0.467 & 0.208   &   3.4              \\ \cline{2-13} 
                        & ours    &  \textbf{0.999}             &    \textbf{0.881}     & 0.142           &0.041       &          2.381  & \textbf{0.188}           &              \textbf{4.06}           & 0.103 & \textbf{0.388}    & \textbf{0.320}  &\textbf{4.2}                 \\ \hline
\end{tabular}}
\label{tab:r2}
\end{table*}

\begin{table*}[t]
\caption{Comparison with different fusion modules for human motion generation on LaserHuman. ``w/o q'' is the method without the queried-cross attention strategy.}
\vspace{-3mm}
\resizebox{\linewidth}{!}{
\vspace{-1ex}
\begin{tabular}{c|ccccccccccc}
\hline
\rowcolor{blue!5}              Methods &non-collision$\uparrow$ & contact$\uparrow$ & APD (t)$\uparrow$ & std (t)$\uparrow$ & APD (p)$\uparrow$ & std (p) $\uparrow$
              & APD (m)$\uparrow$ & std (m)$\uparrow$
              & FID $\downarrow$& R-score$\uparrow$  & p-score $\uparrow$ \\ \hline
w/o qf         &   \textbf{0.999}           & 0.469  &   0.726 &  0.204     & 12.145  &  0.526         &  3.974 & 0.103 & 2.191                   & 0.144                       & 3.3                                 \\ \hline
Triple-Queried & 0.996                             & 0.476                       & \textbf{2.332}                          & 0.660                          & \textbf{12.920}                         & 0.587                          & \textbf{4.172}                           & 0.109                           & 4.120                   & 0.248                       & 2.9                                 \\ \hline
Scene-Queried\cite{vuong2023language}   & \textbf{0.999}                             & \textbf{0.596}                       & 0.974                          & 0.276                          & 13.164                         & 0.605                          & 4.171                           & 0.110                           & 4.063                   & 0.219                       & 3.6                                 \\ \hline
Text-Queried   &  0.994                                 &   0.449                          &   0.609                             &  0.171                              &  12.296                              &   0.539                             &  4.006                               &   0.104                              & 2.782                   & \textbf{0.363}                       & 3.7                                 \\ \hline
ours           & \textbf{0.999}                             & 0.523                       & 2.035                          &\textbf{ 0.830 }                         & 12.903                         & \textbf{0.847}                          & 4.117                           & \textbf{0.152}                           & \textbf{0.987}                   & 0.326                       & \textbf{4.1}                                 \\ \hline
\end{tabular}
}
\label{tab:ablation}
\vspace{-2ex}
\end{table*}
\subsection{Ablation and Comparison for Multi-condition Fusion Method}
To verify the effectiveness of our multi-condition fusion methods, we compare it with various fusion strategies in Table.~\ref{tab:ablation}.

\noindent\textbf{Scene-Queried Dual Fusion}
This is to apply a dual scene-queried cross attention as in \cite{vuong2023language}. The scene feature will be used as query twice and the keys, values are from language feature and enhanced scene feature $F_{pq}$.
\begin{equation}
\begin{aligned}
F_{pq} &= LN(FFN(\text{CA}(F_{pc}', F_l', F_{l}') + F_{pc}'),\\
F_P &= LN(FFN(\text{CA}(F_{pc}', F_{pq}, F_{pq})+F_{pc}').
\end{aligned}
\end{equation}
Finally, $F_P$ is concatenated with language feature $F_l'$ to get multi-modal condition $z_c$.
The strategy overlooks the scene condition and neglects the text instruction. 
The result indicate relatively high FID and low R-score, which has low consistency with text.

\noindent\textbf{Text-Queried Dual Fusion}
This is similar to Scene-Queried approach but replace the scene queries with text queries. This approach lacks fine-grained information on text descriptions for local motion instruction. Additionally, the redundancy in scene information necessitates language-based guidance to focus on interaction-specific areas. It lacks the interaction from text to scene. 

\noindent\textbf{Triple Fusion}
To valid the efficacy of parallel cross-attention mechanism, we implement a triple-fusion method, which retains the text-queried cross-attention and simplifies scene-queried path to a single-layer weighted map. 
We compute the similarity map by $W = \text{Softmax}(F_{pc}'^TF_l')$ and multiply it with $F_{pc}'$ to get enhanced $F_P' = (W^TF_{pc})$. The final fusion condition is the concatenation of three parts, $z_c = FC(F_P'\oplus F_l'\oplus F_L)$.
However, this method encounters a domain gap between language and point features, lacking effective cross-modal mapping.

\noindent\textbf{Parallel Cross Fusion}
Our method efficiently integrates the action and motion direction knowledge from textual features and dynamic scene information from point clouds. This fusion of modalities mutually enhances feature representation, guiding the model toward generating results that are both consistent with textual descriptions and plausible within the scene context.

\subsection{Discussion}
\begin{figure}[t]
    \centering
\includegraphics[width=1.00\columnwidth]{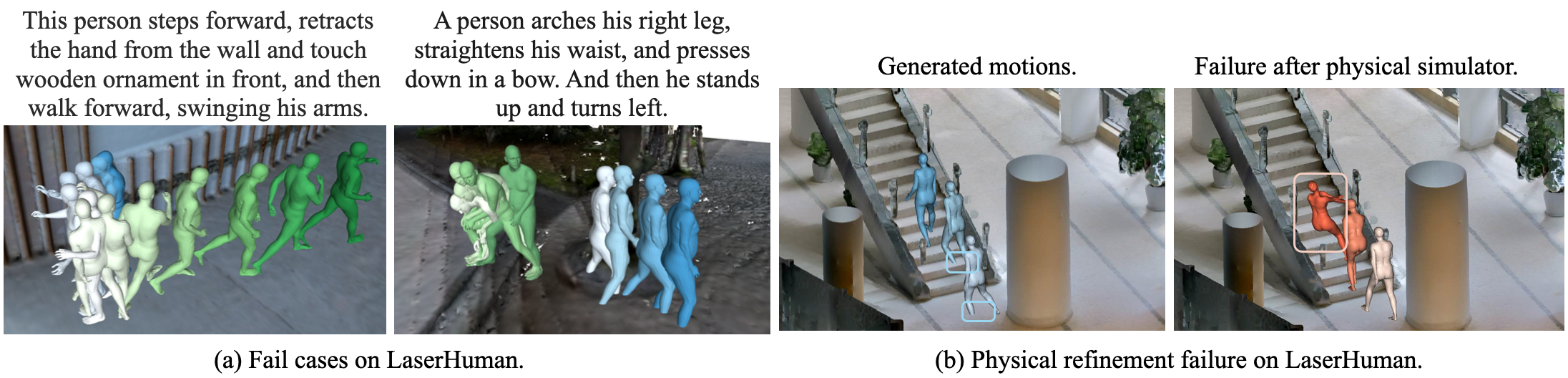}
    \vspace{-4ex}
\caption{Fail cases and physical refinement failure on LaserHuman. In (a), the green person represents reference motions. In (b), the left is the result with penetration and the right is the result after physical simulator.}
    \label{fig:fail}
    \vspace{-4ex}
\end{figure}

\noindent\textbf{Discussion of failure cases.}
We provide some typical examples of failure cases in 
In terms of the consistency of the text description,
the language descriptions are long and challenging with fine-grained details, our results partly align with lengthy text descriptions, focusing mainly on several actions. In addition, the model struggles with complex movements such as  ``arch'', mainly because of the data distribution imbalances. The model will converge on several common cases and have poor performance on less common actions. 
From the perspective of scene-plausible, it remains difficulties dealing with complex terrains such as stairs and irregular ground surfaces, penetration, and floating still exist in our results, as shown in Fig. ~\ref{fig:fail}. In scenarios involving dynamic interactions, the generated motions often lack precision in point-level contact placement. Future research could explore the integration of constraints for both physical and dynamic interactions, alongside the development of an efficient feature extraction method for diverse and dynamic environments.

\noindent\textbf{Physical refinement.} In order to improve physical plausibility of scene-aware motion generation, some works consider applying physical constraints in the post-process or in the diffusion process. Built upon physics simulation environments~\cite{makoviychuk2021isaac} and Deep Reinforcement Learning (DRL), human motion imitator could various motor skills~\cite{peng2018deepmimic,dou2023c, yuan2020residual, wang2023learning,luo2023perpetual,luo2023universal}.
It has been applied to model physically plausible human dynamics during motion generation\cite{yuan2023physdiff}.

We experiment with incorporating physical constraints through a physics-based tracker to generate a physically plausible human motion while encountering several challenges. 
We start by pre-training our model on the AMASS dataset~\cite{mahmood2019amass} following ~\cite{luo2023perpetual}, and then finetune the model using the collected data with captured scenes imported for simulation. 
We illustrate a typical failure in Fig. \ref{fig:fail}. Even fine-tuned on the new dataset, the tracker still fails when going upstairs and results in a person falling over. We attribute the reasons as follows: 
Firstly, existing physical simulation environments~\cite{isaacgym,todorov2012mujoco} typically demand the use of imported clean meshes and are sensitive to noise.
The indoor capture system ensures relatively precise motion detection. However, our result is relatively noisy in contrast to data obtained from mocap systems. 
Thus, utilizing these relatively noisy meshes poses a challenge in tracking motion with a physical simulator where the character's movement can be affected by undesired noisy geometry primitives. 
Secondly, the captured human motions are mainly on flat ground with no changing landforms involved. Yet the dynamic world contains moving and changing objects on complex landforms. Directly applying the physical tracker to our dataset with minimal fine-tuning on finite data of each land case, yielding unsatisfying results. 
In the future, a more effective curriculum learning can be implemented by developing a more robust control policy as the motion imitator trained on a clean motion dataset~\cite{mahmood2019amass} to motion data with varying degrees of noise, incorporating scene complexities in a gradual progression from easy to challenging samples. 

\noindent\textbf{Diverse modalities for human motion generation.}
We provide various types of scene data to facilitate the research on both static and dynamic scene conditional motion generation, as introduced in Section.~\ref{sec:statistics}. Our benchmark is built on the first kind, where dynamic scenes are represented by point clouds. It is interesting to decouple the static scene map and dynamic objects, using the SMPL models for interactive participants.
Exploring the integration of the diverse modalities and extracting both static and dynamic information has significant potential for motion generation, which further benefit animation and the realm of humanoid robots. 

\section{Conclusion}
We introduce LaserHuman, a Scene-Text-to-Motion dataset containing diverse human motions, free-form language descriptions, and unconstrained real-life scenarios. We provide multi-modal data to facilitate the research for conditional motion generation. To generate semantically consistent and physically plausible human motions, we develop a multi-conditional diffusion model and achieve state-of-the-art performance on collected dataset and open dataset. 

\clearpage  

%
%
\bibliographystyle{splncs04}
\bibliography{main}

\appendix
\clearpage
\noindent \textbf{\Large Appendix}\\

We present additional statistics of LaserHuman and more examples of motions generation results.
\appendix
\section{Dataset}
\label{sec:sup_dataset}
\subsection{Statistics of LaserHuman}
\begin{figure*}
   \centering
\includegraphics[width=1.0\columnwidth]{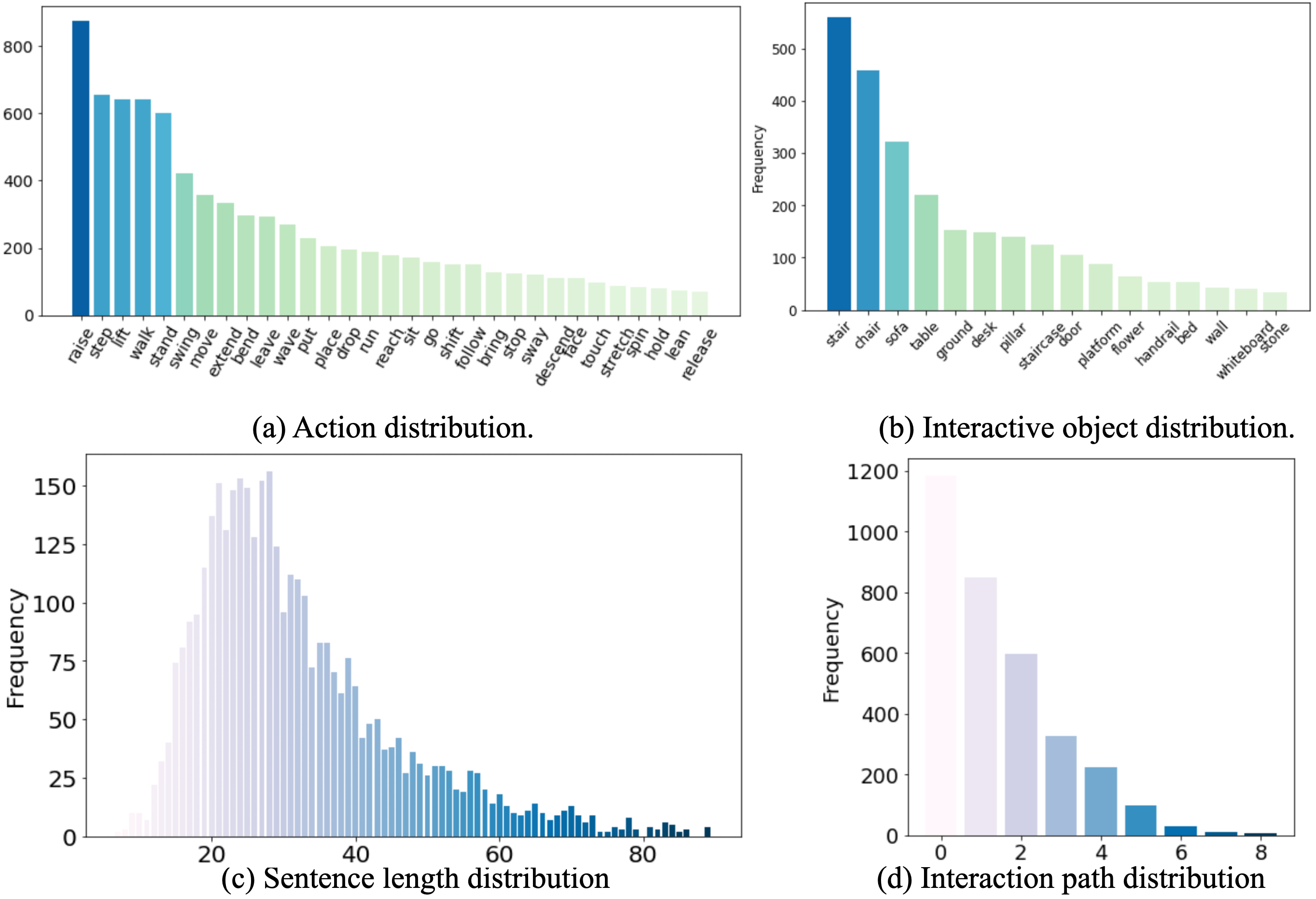}
   \caption{Statistics of the LaserHuman dataset.}
   \label{fig:sup_statistics}
\end{figure*}

We present additional statistics of the LaserHuman dataset. The distribution of actions and interactive objects is depicted in Fig.\ref{fig:sup_statistics} (a) and (b). We identify the 30 most frequent verb categories. During the statistical analysis, certain position verbs, such as `turn, do' were excluded. It is noted that a single description may contain two or more verbs. Given that we include complex terrains in both outdoor and indoor scenes, the most commonly encountered objects are `stairs,' `chair,' and `sofa.' The average length of the annotated language descriptions is 28.5 words. The distribution of sentence lengths in these language descriptions is shown in Fig.\ref{fig:sup_statistics} (c). Fig.\ref{fig:sup_statistics} (d) illustrates the distribution of interaction paths, where we calculate the path length of interactions within the scene. For annotation purposes, we set a maximum duration of 4 seconds and a maximum distance of 9 meters between the human and the environment.
\subsection{Processing details}
The synchronization between IMUs
and the LiDAR is by detecting the jumping peak between multiple sensors and the synchronized frequency is 10Hz.
We include interacted persons in the scenes, and the SMPL models of these interacting participants are obtained using LIP~\cite{ren2023lidar} and further optimized from multi-view LiDAR point clouds. Specifically, we process the raw point cloud to extract the instance point cloud sequences of each person from three different views generate the SMPL pose parameters. The LiDAR-branch of LIP has been modified with a multi-extractor to efficiently extract feature from human point cloud sequences obtained from these three different views. We utilize Chamfer Distance (CD) loss to select the best match with the point cloud.

\subsection{Additional motion sequences cases.}
\begin{figure*}
   \centering
\includegraphics[width=1.0\columnwidth]{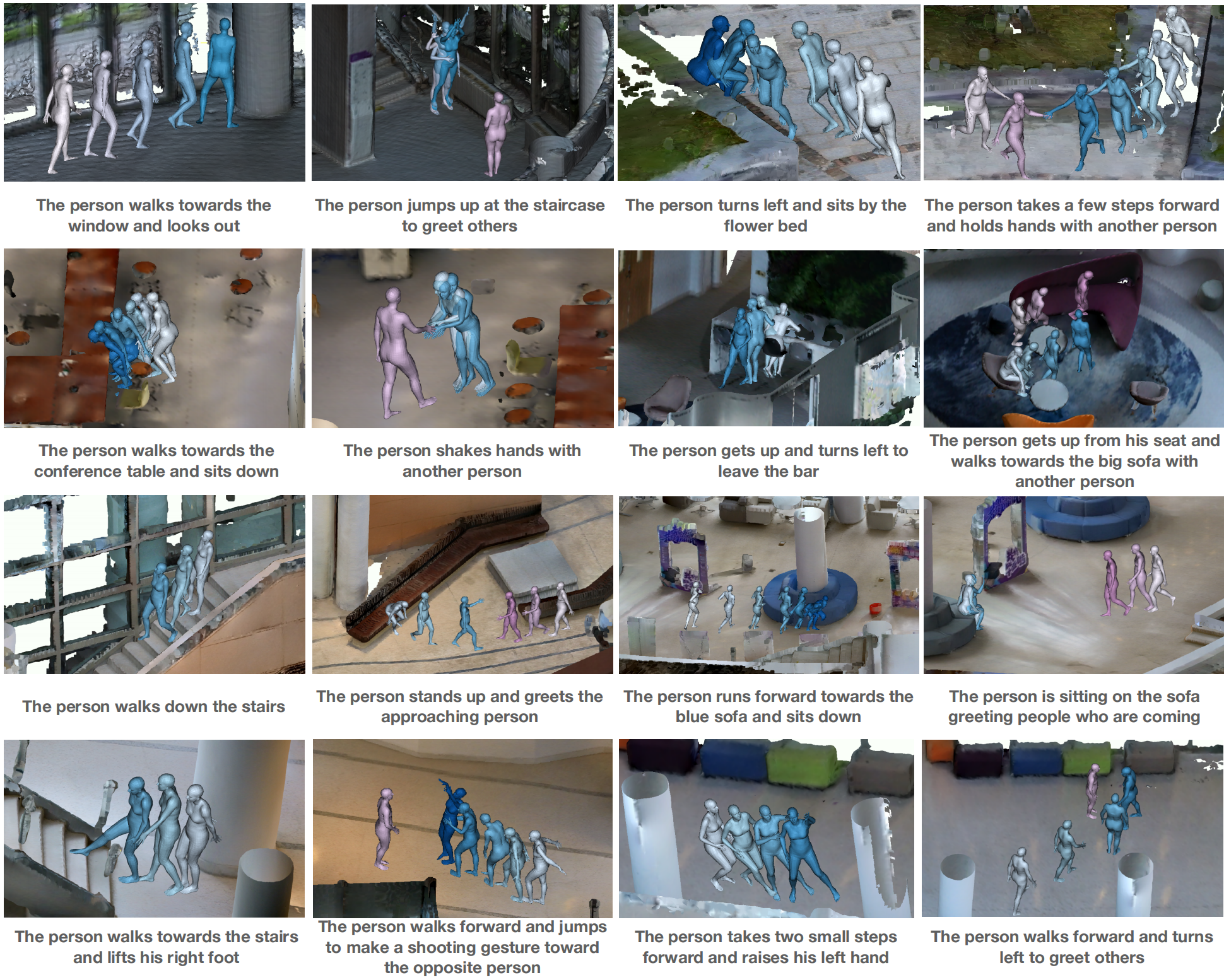}
   \caption{Additional motion sequences cases.}
   \label{fig:sup_gallery}
\end{figure*}

We provide more motion sequences cases in Fig.\ref{fig:sup_gallery}.

\section{Implementation Details}
Following prior work ~\cite{mdm2022human} as our baseline, we further designed the multi-condition fusion model.  
The motion sequences are clipped or padded to a fixed duration of 40 frames (4 seconds). The point cloud input consists of the cropped scene map and the dynamic interactive person segmented from the LiDAR point cloud, and this concatenated result is then downsampled to $N_p=32768$, providing raw information about the 3D scene.
We use the Adam optimizer with a learning rate of 0.0001. The batch size is set to 64, and the number of diffusion steps, T, is set to 1000.
\section{Additional Results}

\subsection{Additional ablation studies}
To verify the effectiveness of our multi-condition fusion methods, we compare it with various fusion strategies. We implement our designed parallel cross fusion module on SceneDiff\cite{scenediff} and denotes this as sd-Text(w cf). The improved results in Table. ~\ref{tab:ablation-sup} demonstrate the efficacy of our fusion module.
\subsection{Additional qualitative results}
\begin{table*}[ht]
\caption{Comparison with different fusion modules for human motion generation. ``w/o q" is the method without the queried-cross attention strategy and ``w cf" is implementing our cross fusion on sd-Text .}
\vspace{-3mm}
\resizebox{\linewidth}{!}{
\vspace{-1ex}
\begin{tabular}{c|ccccccccccc}
\hline
\rowcolor{blue!5}              Methods &non-collision$\uparrow$ & contact$\uparrow$ & APD (t)$\uparrow$ & std (t)$\uparrow$ & APD (p)$\uparrow$ & std (p) $\uparrow$
              & APD (m)$\uparrow$ & std (m)$\uparrow$
              & FID $\downarrow$& R-score$\uparrow$  & p-score $\uparrow$ \\ \hline
w/o qf         &   \textbf{0.999}           & 0.455  &   0.674 &  0.232     & 2.704  &  0.105         &  2.557 & 0.097 & 2.010                   & 0.210                       & 3.5                                 \\ \hline
Triple-Queried & 0.999         & 0.369   & 0.622        & 0.217        & 2.712       & 0.115    &2.968 & 0.116                   & 1.949 & 0.198      &3.0                  \\ \hline
Scene-Queried\cite{vuong2023language}  & 0.999         & 0.410  & 0.740        & 0.173        & 2.525     &  0.073    &2.767 & 0.072   &  2.322 & \textbf{0.212} & 3.1                               \\ \hline
Text-Queried        & 0.999         & 0.376  & 0.646        &  0.225        & 2.718     &  0.109    &2.663 &  0.101   &  
\textbf{1.818} & 0.208   & 3.3                                          \\ \hline
sd-Text(w cf)           & \textbf{0.999}                             & \textbf{0.472}                       &\textbf{0.772}                          &\textbf{ 0.273 }                         & \textbf{2.726}                         & \textbf{0.126}                          & \textbf{3.203}                           & \textbf{0.126}                           &         1.859 & 0.196          & 3.7                                \\ \hline
\end{tabular}
}
\label{tab:ablation-sup}
\vspace{-2ex}
\end{table*}

\begin{figure*}
   \centering
\includegraphics[width=1.0\columnwidth]{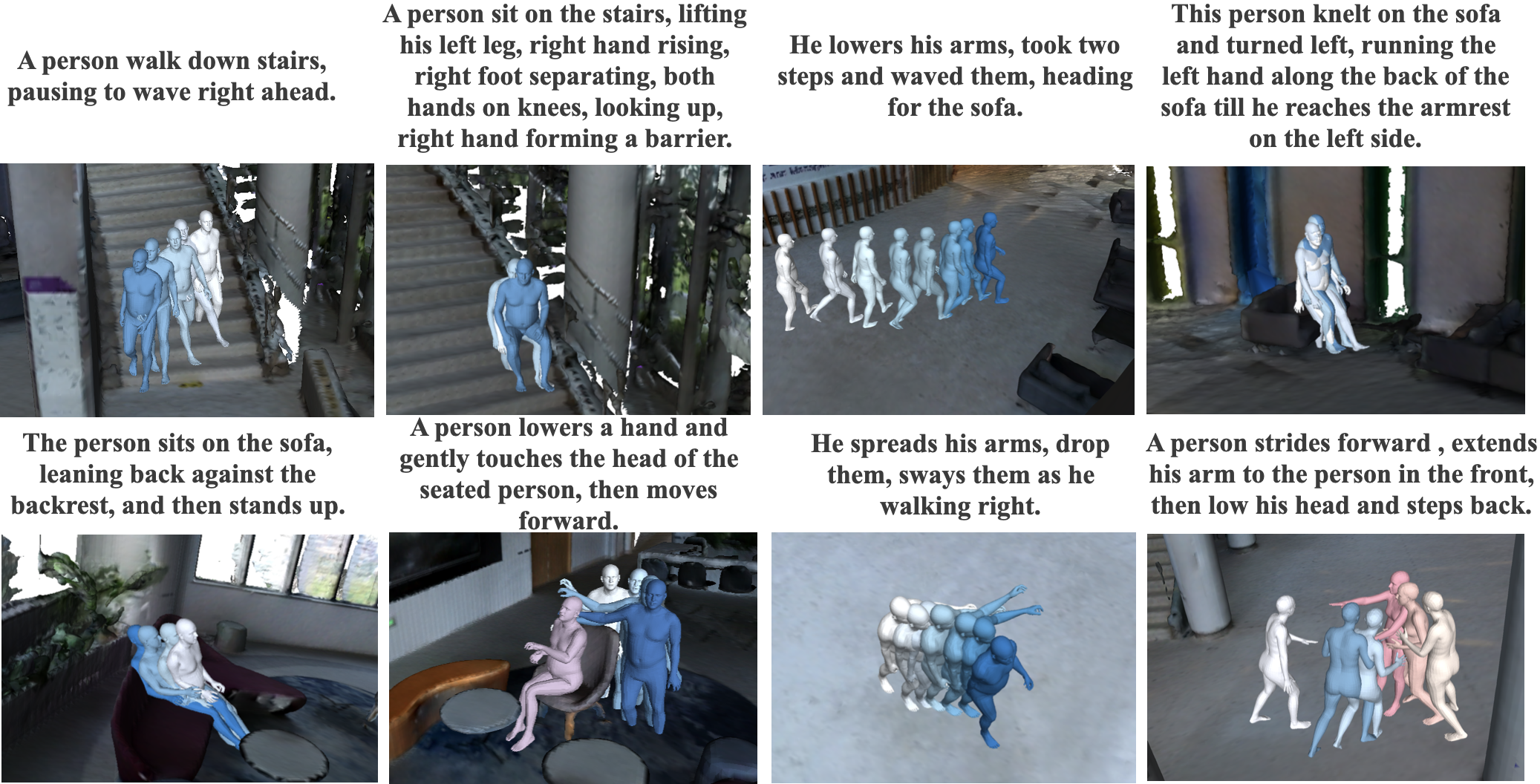}
   \caption{Motion generation results on  LaserHuman. The persons in pink are the interacted participants.}
   \label{fig:vis-more}
\end{figure*}

More motion generation results on LaserHuman are shown in Fig. ~\ref{fig:vis-more}.

\clearpage  

%
%

\end{document}